\documentclass[sigconf]{acmart}
\AtBeginDocument{%
  }
\usepackage{algorithm}
\usepackage{algorithmic}
\setcopyright{rightsretained}

\usepackage{multirow}
\usepackage{subcaption}

\begin{document}

\title{Active Learning for Multiple Change Point Detection in Non-stationary Time Series with Deep Gaussian Processes}


\author{Hao Zhao}
\affiliation{%
  \institution{Arizona State University}
  \city{Tempe}
  \state{Arizona}
  \country{USA}
}
\email{hzhao83@asu.edu}

\author{Rong Pan}
\affiliation{%
  \institution{Arizona State University}
  \city{Tempe}
  \state{Arizona}
  \country{USA}
}
\email{Rong.Pan@asu.edu}

\renewcommand{\shorttitle}{AL for MCP Detection in Non-stationary Time Series with DGPs}

\begin{abstract}

Multiple change point (MCP) detection in non-stationary time series is challenging due to the variety of underlying patterns. To address these challenges, we propose a novel algorithm that integrates Active Learning (AL) with Deep Gaussian Processes (DGPs) for robust MCP detection. Our method leverages spectral analysis to identify potential changes and employs AL to strategically select new sampling points for improved efficiency. By incorporating the modeling flexibility of DGPs with the change-identification capabilities of spectral methods, our approach adapts to diverse spectral change behaviors and effectively localizes multiple change points. Experiments on both simulated and real-world data demonstrate that our method outperforms existing techniques in terms of detection accuracy and sampling efficiency for non-stationary time series.

\end{abstract}

\ccsdesc[500]{Computing methodologies~Gaussian processes}

\keywords{Change Point Detection, Deep Gaussian Process, Active Learning}


\maketitle

\section{INTRODUCTION}
\label{Sec:INTRO}

\begin{figure*}[t]
  \centering
  \begin{subfigure}[b]{0.19\linewidth}
    \includegraphics[width=\linewidth]{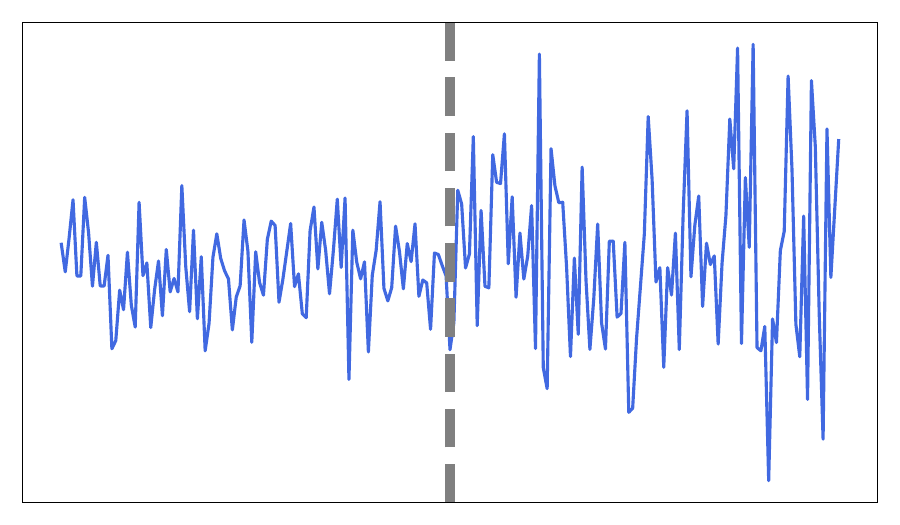}
     \caption{Stratified Change}
  \end{subfigure}
  \begin{subfigure}[b]{0.19\linewidth}
    \includegraphics[width=\linewidth]{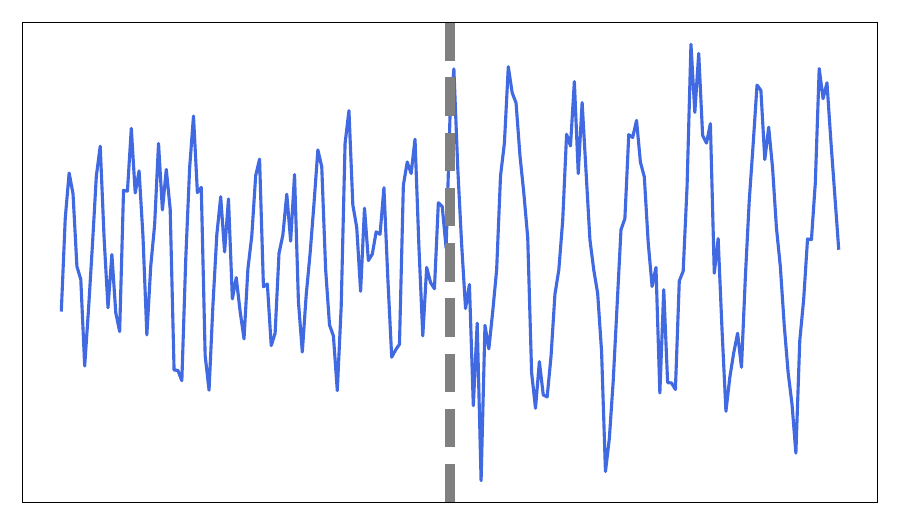}
    \caption{Cycle Change}
  \end{subfigure}
  \begin{subfigure}[b]{0.19\linewidth}
    \includegraphics[width=\linewidth]{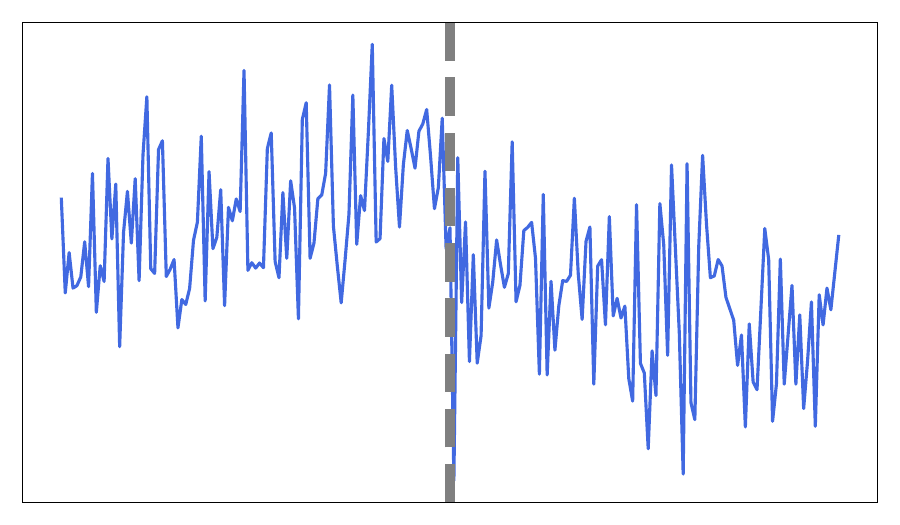}
    \caption{Trend Change}
  \end{subfigure}
  \begin{subfigure}[b]{0.19\linewidth}
    \includegraphics[width=\linewidth]{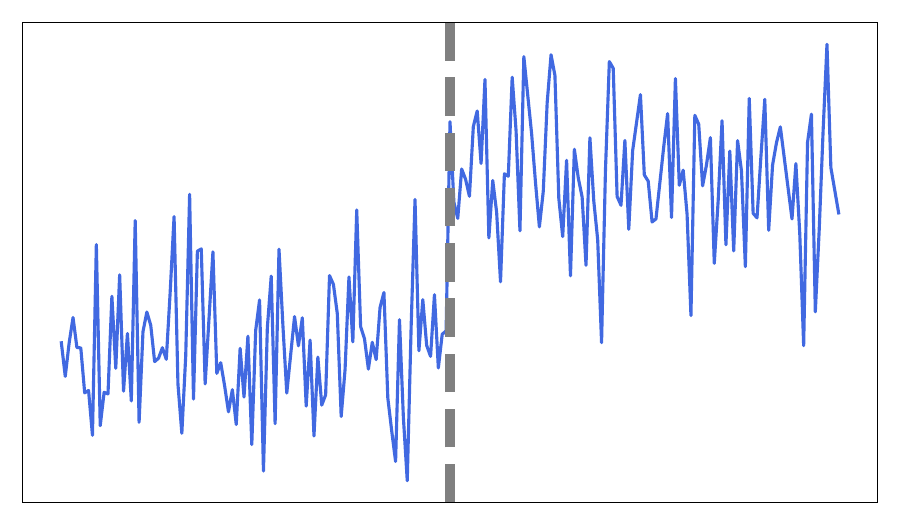}
    \caption{Shift Change}
  \end{subfigure}
  \begin{subfigure}[b]{0.19\linewidth}
    \includegraphics[width=\linewidth]{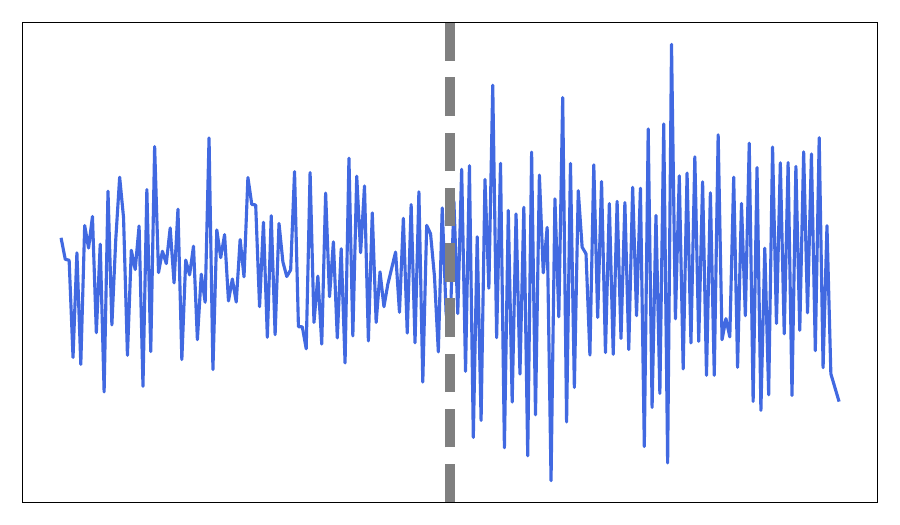}
    \caption{Systematic Change}
  \end{subfigure}
  \caption{Examples of Change Patterns in Stochastic Time Series. Each subplot illustrates a unique structural change occurring at the middle, denoted by a grey dashed vertical line.}
  \Description{Examples of change patterns.}
  \label{Fig:Change Patterns}
\end{figure*}

Change point detection (CPD) deals with identifying structure changes occurring in a sequence of observed samples from a temporal or spatial stochastic process. This task is applicable across a wide variety of domains, including material transition phenomena \citep{hayashi2019active}, environmental monitoring \citep{ oelsmann2022bayesian}, industrial process control \citep{schat2023detecting}, and financial forecasting \citep{casini2024change}, etc. CPD methods are typically categorized into online detection methods, which detect changes as they occur, and offline detection methods, which retrospectively identify changes after receiving all samples. Here, our work focuses on offline detection, while assuming assessing these samples can be very costly.

In certain scenarios, acquiring useful data for CPD can be very expensive, time-consuming, and resource-intensive. Examples include phase transitions in material science \citep{schiepek2020convergent}, seafloor-depth assessment \citep{toodesh2021prediction}, atmospheric and water pollution monitoring \citep{ropkins2021early, shen2015impact}, and drilling operations in oil and mining \citep{grzesiek2021method}. These tasks often require intricate and accurate measurements, and it beneficial to strategically select measurement conditions or locations rather than relying on random or exhaustive sampling. Active Learning (AL) is well-suited for this purpose, as it enables the selection of the most informative data points, reducing the cost of data acquisition without compromising CPD performance.

To address the limitations of existing CPD methods, particularly regarding the diversity of possible change patterns in non-stationary time series (see Figure~\ref{Fig:Change Patterns} for the examples), we propose a novel algorithm that integrates Deep Gaussian Processes (DGPs) \citep{damianou2013deep, sauer2023vecchia} with spectral analysis and an AL framework. Our methodology begins by employing DGPs as predictive models, which not only filter out noise but also adapt to the non-stationary characteristics of the data. The mean predictions generated by the DGPs are subsequently transformed into spectral representations using a sliding window Fourier Transform. This transformation extracts localized frequency-domain features essential for detecting spectral changes. Spectral uncertainty is quantified by Monte Carlo (MC) sampling on these spectral representations. This uncertainty estimation is then incorporated into an Acquisition Function (AF) that balances two core objectives in AL -- exploitation and exploration. Exploitation is driven by identifying spectral regions that likely contain changes based on the Spectral Change Detection Metric (SCDM), while exploration targets areas where the spectral uncertainty is high, suggesting further data acquisition in less understood regions of the time series. By iteratively selecting data points guided by this AF, our algorithm refines its predictive accuracy and focuses measurement effort on the most relevant portions of the input space. We validate our method using both synthetic and real-world experiments, and it demonstrates robust performance across a range of complex change patterns. The results indicate that this combined DGP–Spectral–AL framework can flexibly adapt to a wide variety of changing dynamics and noise levels, which makes it well-suited for problems in complex, non-stationary environments.\\

The main contributions of this paper are:
\vspace{-5pt}
\begin{itemize} 
\item We deploy DGPs as flexible, non-parametric models to model non-stationary time series CPD. This approach accurately captures the complex temporal/spatial dependencies and non-linear relationships within the data. 

\item We develop a Spectral Change Detection Metric (SCDM) that integrating spectral mean change and weighted spectral gradient distance. The SCDM enables robust detection of a wide range of change behaviors in the frequency domain. 

\item We propose an AF within the AL framework that optimizes the sampling process by balancing exploration and exploitation. The AF integrates the SCDM with spectral uncertainty estimates obtained via MC sampling, and guides the selection of the most informative data points to enhance CPD performance while minimizing data acquisition costs. 
\end{itemize}

The remainder of this paper is organized as follows: Section~\ref{Sec:RW} reviews the related topics, Section~\ref{Sec:BACKGROUND} provides the necessary background on GPs and DGPs. The proposed approach is thoroughly explained in Section~\ref{Sec:PROPOSED METHOD}. Section~\ref{Sec:EXPERIMENTS} describes the experimental setup and presents the results. Finally, Section~\ref{Sec:CONCLUSION} concludes the paper.

\section{RELATED WORK}
\label{Sec:RW}

\paragraph{CPD} 
Classical CPD methods primarily detect simple statistical patterns such as shifts in process mean. Traditional methods include Statistical Process Control (SPC) charts, such as the Exponentially Weighted Moving Average (EWMA) Chart \citep{sukparungsee2020exponentially}, Cumulative Sum (CUSUM) Chart \citep{yu2022comprehensive}, and Moving Average (MA) Chart, are widely used to detect mean changes by testing hypotheses about the constancy of underlying statistical properties before and after a potential change point \citep{pan2012bayesian, steward2016bayesian}. These models typically assume process stationarity, and their performance heavily depends on whether the actual data follows the predefined distribution. 

In contrast, Machine learning (ML) based time-series analysis provides a broader framework for CPD by relaxing strict parametric constraints. Advanced models such as neural networks \citep{zhang2020feature}, GPs \citep{caldarelli2022adaptive}, and deep learning architectures \citep{gupta2022real} are capable of capturing complex temporal dynamics and nonlinear relationships. This flexibility enables ML-based methods to effectively detect change points in dynamic environments and handle diverse change patterns that traditional statistical methods might overlook.

\paragraph{CPD in Frequency Domain} 
CPD has also been explored in the frequency domain across multiple fields \citep{casini2024change}. Spectral methods typically involve comparing the spectral representations of different segments to identify significant changes. Common spectral distance metrics such as Euclidean distance, Spectral Angle Mapper (SAM), Spectral Correlation Mapper (SCM), and Spectral Gradient Difference (SGD) have been widely employed to quantify discrepancies in spectral shapes, particularly within land cover change studies \citep{heydari2023remote, zhang2022continuous, yan2018novel}. Additionally, \cite{zhang2023spectral} introduced a Spectral CUSUM distance-based method for detecting network structural changes, including the emergence of new communities and switching memberships. Furthermore, several researchers have investigated methods based on segmenting the wavelet spectrum for piecewise stationary time series \citep{cho2015multiple, korkas2017multiple}. A key advantage of frequency-domain methods is their ability to detect changes without making explicit assumptions about the data-generating mechanism. Unlike time-domain detection, which typically focuses on one or two specific change patterns, spectral methods offer a more comprehensive and robust change detection capability. 

\paragraph{GPs-based CPD} 
GPs and DGPs \citep{damianou2013deep, sauer2023vecchia} are well-suited for AL in regression because they provide principled uncertainty quantification \citep{sauer2023active}. Recent GP-based CPD approaches include the Derivative-Aware Change Detection (DACD) method \citep{zhao2023active}, which employs GP derivatives based AL to guide data queries around likely locations of change points; and the work by \cite{booth2023contour}, which integrates DGPs into contour location for aerospace simulations by utilizing a hybrid criterion that explores along the Pareto front of entropy and uncertainty to detect changes. However, these methods are designed to handle different types of changes separately (e.g., mean shifts, variance changes), which limits their applicability in more complicated scenarios where multiple types of change patterns exist. By contrast, our proposed method integrates spectral analysis with a flexible DGP framework and a custom AL strategy to accommodate a broader range of changes in non-stationary time series.

\section{Background}
\label{Sec:BACKGROUND}

\subsection{Change Point Detection}
Offline CPD identifies the locations in a data sequence where the sequence's statistical properties exhibit significant changes. These changes could involve various characteristics such as the mean, variance, correlation structure, or even the entire distribution of the data. Suppose there are $N$ observations $\mathcal{D} = (X, Y)=\{(x_i, y_i)\}^N_{i=1}$, where the input $x_i \in \mathcal{X}$ (i.e., time stamps) and the output $y_i \in \mathcal{R}$ (i.e., system responses) follow an unknown function $f(y_i|x_i)$. In the case of a time series, the objective of the CPD algorithm is to identify the change times $x^{\text{cp}}_{k}\in\mathcal{X}$ from a collection of $T$ time series denoted as $\{x_1, \cdots, x_T\}$. The subscript $k\in\{1,\cdots, K\}$, where integer $K$ corresponds to the total number of changes in the time series. Thus, the whole time series consists of $K+1$ segments and the underlying statistical property of each segment is different from its adjacent segments.

\subsection{Gaussian Processes}
A GP is a stochastic process comprised of random variables indexed through either time or space, and each finite group of these variables follows a joint Gaussian distribution. Given a dataset $\mathcal{D} = {(\mathbf{x}_i, y_i)}_{i=1}^N$, where $\mathbf{x}_i \in \mathbb{R}^D$, and $y_i \in \mathbb{R}$ with $\mathbf{X} = (\mathbf{x}_1, \dots ,\mathbf{x}_N)^\top$, $ \quad \mathbf{y} = (y_1, \dots, y_N)^\top$. GP defines a prior distribution over functions $\mathbf{f}=f(\mathbf{X})$ with a mean function $m:\mathbb{R}^D \rightarrow \mathbb{R}$ and a covariance function $k:\mathbb{R}^D \times \mathbb{R}^D \rightarrow \mathbb{R}$. The GP posterior can yield an analytically tractable predictive distribution for any test point $\mathbf{x}^\star$. The inference step requires inverting an $N \times N$ covariance matrix, which has a complexity of $\mathcal{O}(N^3)$. Thus, GPs are computationally expensive for large datasets. Approximations such as Sparse Gaussian Processes (SGPs) or inducing points are often employed to make them more scalable.

SGPs intruduce a set of $M$ inducing points $\mathbf{Z} =(\mathbf{z}_1,\cdots,\mathbf{z}_M)^T$ and the corresponding function values $\mathbf{u} = f(\mathbf{Z})$ with $p(\mathbf{u}) = \mathcal{N}(\mathbf{u} | m(\mathbf{Z}), k(\mathbf{Z}, \mathbf{Z}))$. The joint density of $\mathbf{y}, \mathbf{f}, \mathbf{u}$ is
\begin{equation}
p(\mathbf{y}, \mathbf{f}, \mathbf{u}) = p(\mathbf{f} | \mathbf{u}; \mathbf{X}, \mathbf{Z}) p(\mathbf{u}; \mathbf{Z}) \prod_{i=1}^{N} p(\mathbf{y}_i | \mathbf{f}_i).
\end{equation}

With the variational inference, the posterior $p(\mathbf{f},\mathbf{u}|y)$ is approximated by a variational distribution $q(\mathbf{f}, \mathbf{u}) = p(\mathbf{f} | \mathbf{u}; \mathbf{X}, \mathbf{Z}) q(\mathbf{u})$ where $q(\mathbf{u}) = \mathcal{N}(\mathbf{u} | \mathbf{m}, \mathbf{S})$. The Evidence Lower Bound (ELBO) is
\begin{equation}
\mathcal{L}_{SGP} = \mathbb{E}_{q(\mathbf{f}, \mathbf{u})} \left[ \log \frac{p(\mathbf{y}, \mathbf{f}, \mathbf{u})}{q(\mathbf{f}, \mathbf{u})} \right],
\end{equation}
which can be optimized to find the variational parameters $ \mathbf{m}, \mathbf{S}$. By marginalizing $\mathbf{u}$, the posterior on $ \mathbf{f} $ is $q(\mathbf{f}) = \mathcal{N}(\mathbf{f} | \tilde{\mathbf{m}}, \tilde{\mathbf{\Sigma}})$,
where
\begin{equation}\label{eq:mean}
    \tilde{\mathbf{m}}_i = \mathbf{m}(\mathbf{x}_i) + \alpha(\mathbf{x}_i)^\top (\mathbf{m} - \mathbf{m}(\mathbf{Z})),
\end{equation}
\begin{equation}\label{eq:var}
\tilde{\mathbf{\Sigma}}_{ij} = k(\mathbf{x}_i, \mathbf{x}_j) - \alpha(\mathbf{x}_i)^\top (k(\mathbf{Z}, \mathbf{Z}) - \mathbf{S}) \alpha(\mathbf{x}_j),    
\end{equation}
and:
\begin{equation}
\alpha(\mathbf{x}_i) = k(\mathbf{Z}, \mathbf{Z})^{-1} k(\mathbf{Z}, \mathbf{x}_i).
\end{equation}

Common choices for the covariance function include the Radial Basis Function (RBF) kernel, which assumes smoothness in the underlying function and provides infinitely differentiable sample paths, and Matérn kernel, which includes a parameter to control smoothness. Both RBF and Matérn are stationary kernels, they only depend on the relative positions of two inputs and not the absolute locations. While GPs are good at modeling continuous functions, they face limitations when applied to non-stationary processes or functions with sharp jumps. This is because the smoothness assumptions embedded in commonly used kernels (like RBF) make it difficult to capture abrupt changes in the underlying function. Although it is possible to introduce discontinuities by modifying the mean function, or by enforcing independence between covariance functions that apply to different regions of the input space \citep{bitzer2023hierarchical}, this approach requires specifying both the number and locations of the discontinuities as a priori. In this case, the model becomes parametric and less flexible. For situations where the number and locations of discontinuities are unknown, the standard GP framework proves insufficient. 



\subsection{Deep Gaussian Processes}
\begin{figure}
    \centering
    \includegraphics[width=0.75\linewidth]{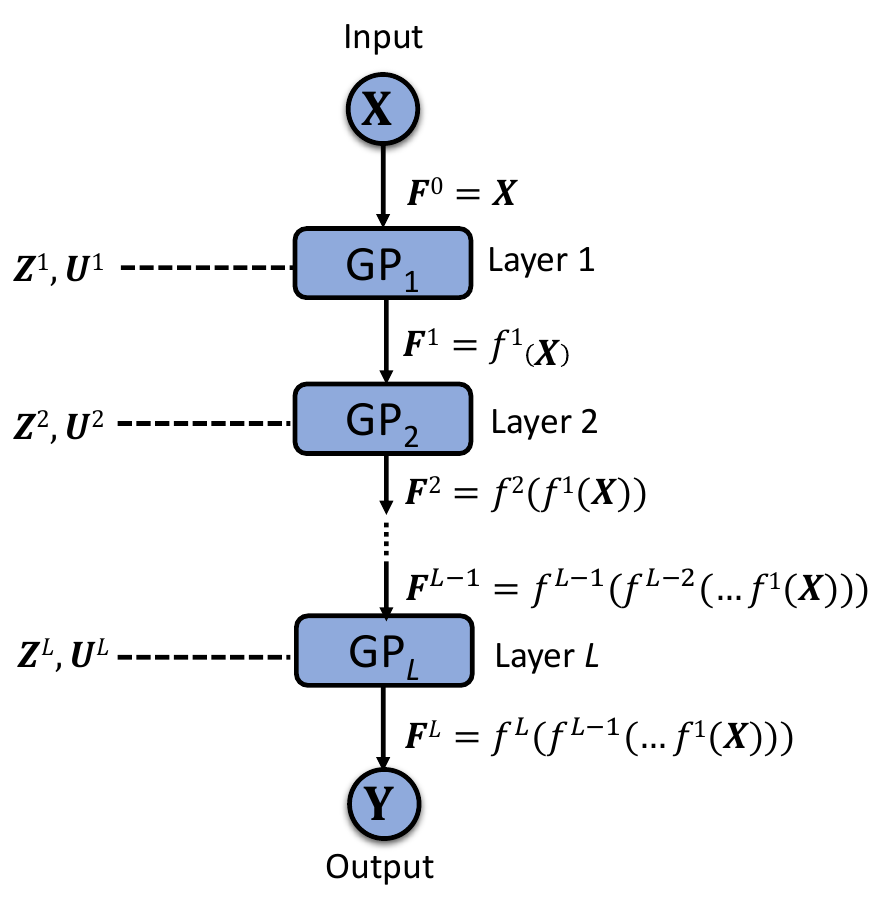}
    \caption{Architecture of a DGP. Each layer $L$ corresponds to a $\text{GP}_l$ with latent function $f^l$ and inducing variables $\mathbf{Z}^l$, $\mathbf{U}^l$. The input to each layer is the output from the previous layer, and the model represents a nested composition $\mathbf{F}^l = f^l(f^{l-1}(\dots f^1(\mathbf{X})))$.}
    \Description{Architecture of a DGP.}
    \label{fig:DGP}
\end{figure}
DGPs extend the power of standard GPs by stacking multiple layers of GPs in a hierarchical structure. 
In a DGP, each layer $F^{l}$ is represented by a GP, and the input of each layer is the output of the previous layer $F^{l-1}$, see Figure~\ref{fig:DGP} for details. Formally, a DGP with $L$ layers can be written as \citep{salimbeni2017doubly}:
\begin{equation}
\begin{aligned}
p(\mathbf{Y}, \{\mathbf{F}^l, &\mathbf{U}^l\}_{l=1}^{L}) = \prod_{i=1}^{N} p(y_i | f_i^L) \times\\
 & \prod_{l=1}^{L} p(\mathbf{F}^l | \mathbf{U}^l; \mathbf{F}^{l-1}, \mathbf{Z}^{l-1}) p(\mathbf{U}^l | \mathbf{Z}^{l-1}),
\end{aligned}
\end{equation}
where $\mathbf{Z}^{l-1}$ denotes the inducing locations at each layer, and $\mathbf{U}^l$ represents the inducing function values. Additionally, $F^0 = X$, which is the input data.

Exact inference is intractable, to address this, we follow the work of \cite{salimbeni2017doubly}. Their algorithm Doubly Stochastic Variational Inference assumes $q(\mathbf{U}^l) = \mathcal{N}(\mathbf{U}^l | \mathbf{m}^l, \mathbf{S}^l)$, then we have $q(\{\mathbf{F}^l\}_{l=1}^L) = \prod_{l=1}^L \mathcal{N}(\mathbf{F}^l | \tilde{\mathbf{\mu}}^l, \tilde{\mathbf{\Sigma}}^l)$, $\tilde{\mathbf{\mu}}^l_i $, and $\tilde{\mathbf{\Sigma}}^l_{ij}$ are defined in Eqs. (\ref{eq:mean}) and (\ref{eq:var}). 

Another assumption is that the posterior distribution of the latent functions is factorizable across layers, with each layer depending only on the previous layer. Formally we have:
\begin{equation}
q(\mathbf{f}_i^L) = \int \prod_{l=1}^{L-1} q(\mathbf{f}_i^l | \mathbf{m}^l, \mathbf{S}^l; \mathbf{f}_i^{l-1}, \mathbf{Z}^{l-1}) \, d\mathbf{f}_i^l.
\end{equation}
This marginal dependence allows efficient recursive sampling across layers using a re-parameterization trick, which simplifies the optimization of the posterior. Then the ELBO is given as
\begin{equation}
\begin{aligned}
\mathcal{L}_{DGP} = &\sum_{i=1}^{N} \mathbb{E}_{q(f_i^L)} [\log p(y_n | f_n^L)] - \\
& \sum_{l=1}^{L} \text{KL}\left[q(U^l) \| p(U^l | \mathbf{Z}^{l-1})\right].
\end{aligned}\end{equation}
In fact, the composition of these GP layers leads to non-Gaussian models, as the final output no longer strictly follows a Gaussian distribution. This flexibility allows DGPs to capture a wider variety of data patterns, especially in non-stationary time series.

While DGPs provide greater expressiveness than standard GPs, they come at a higher computational cost. For a DGP with $L$ layers and $M$ inducing points per layer, the time complexity of inference is approximately $\mathcal{O}(L M^2 N)$, and the memory complexity scales as $\mathcal{O}(L M^2)$. Empirical studies \citep{dunlop2018deep} suggest that DGPs with two or three layers are often sufficient for capturing complex patterns in real-world data without significant computational overhead. Deeper architectures may lead to diminishing returns in accuracy while greatly increasing computational demands.

\section{Proposed Method}
\label{Sec:PROPOSED METHOD}
\begin{figure*}[t]
\centering
\includegraphics[width=\textwidth]{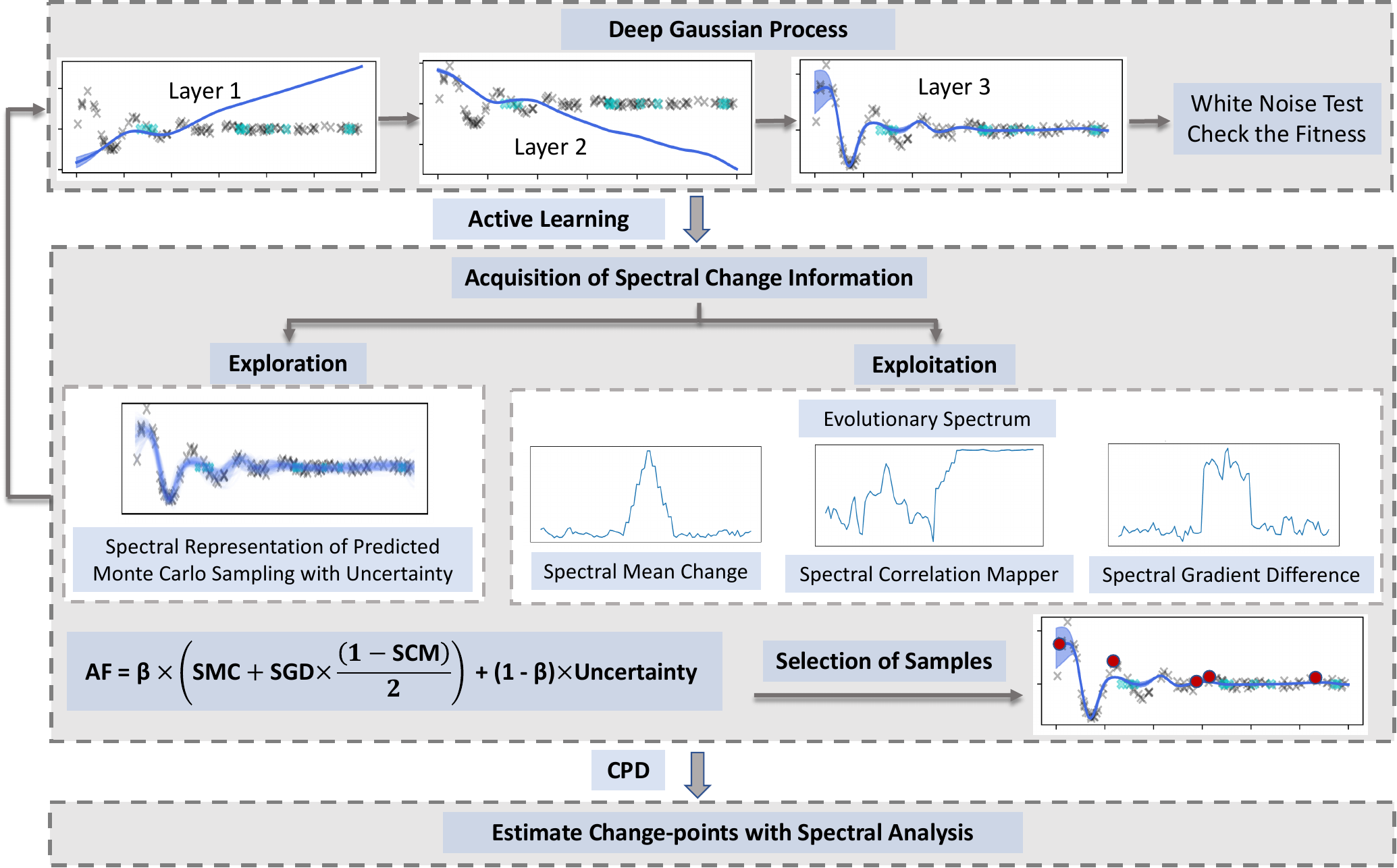}
\caption{Flowchart of Proposed Method. The methodology is achieved via three main steps:
(1) DGP modeling: fit the non-stationary time series data; (2) active sampling: select extra data points based on spectral distance metrics and uncertainty estimates; (3) change point identification: estimate change points through spectral analysis.
}
\Description{Algorithm flowchart.}
\label{Figure:Flowchart}
\end{figure*}

Our methodology is designed for situations where acquiring data is expensive, so we need to make efficient use of a limited initial dataset to model the whole process and detect the changes, as shown in Figure~\ref{Figure:Flowchart}. There are three major steps for our method to obtain the change points: (1) employ DGP to fit the non-stationary time series and to obtain the spectral representation; (2) active sampling additional points based on spectral distance and uncertainty; (3) estimate the change points with spectral analysis.

\subsection{Spectral Representation}
We start with obtaining an overall sketch of the non-stationary time series model by employing a DGP model. In DGP, the original input is spatially warped through composite stationary GP layers to fit the response. This warping results in the "stretch" and "compress" of data in different regions of the input space. The ability to perform these non-linear transformations allows DGP to adapt to various data complexities. Moreover, the hierarchical nature of DGP introduces a localized smoothing effect as each layer refines the predictions of the previous layer, thereby reducing the impact of high-frequency noise.

To ensure the model accurately captures the underlying data structure, we perform a validation test, or the white noise test. We analyze the residuals of the model predictions to check for significant auto-correlation. Passing the white noise test indicates that the residuals are random and the DGP model has effectively modeled the time series.

Once we have the mean value predictions from the DGP, we can perform the spectral analysis by applying a sliding window Fourier Transform (also known as Short-Time Fourier Transform or STFT). It is a widely used technique for analyzing non-stationary signals by breaking the signal into smaller overlapping windows and
analyzing its frequency content over time. This yields the frequency spectrum for that specific time segment as:
\begin{equation}
\label{eq:STFT}
X(\tau, \omega)= \int_{-\infty}^{\infty} x(t) w(t - \tau) e^{-j \omega t} dt,
\end{equation}
where $x(t)$ is the DGP predicted mean, $w(t - \tau)$ is the window function centered at time $\tau$. And we fix the window size and denote it as $A$ for implementation convenience. By moving the window across the time series and repeating the process, we can obtain a time-frequency representation of the data. 

\subsection{New Sampling Point Selection}
To efficiently select new data points for sampling, we design an AF that balances the exploitation and exploration of the underlying data generation mechanism. This function incorporates spectral change metrics to detect significant changes in the time series, and it also utilizes uncertainty estimation via MC sampling to guide the selection of regions where the model is less certain. 

\subsubsection{Spectral Change Metrics}
To quantify changes in spectra, three key spectral change metrics are employed and each of them targets some specific spectral shapes. 

The Spectral Mean Change (SMC) detects changes by comparing the average spectral power across different segments of the time series.
\begin{equation}
\text{SMC}= \sqrt{\sum_{i=1}^{N} \left( \overline{X}_a^i - \overline{X}_b^i\right)^2},
\end{equation}
where $\overline{X}_a$ and $\overline{X}_b$ are the mean of absolute spectra for different segment windows. SMC is ideal for detecting abrupt changes in spectral power, such as shifts in system states or changes in external conditions affecting the signal.

Spectral Correlation Mapper (SCM) computes the difference between spectral shapes by Pearson’s correlation of them with the value range $[-1,1]$.
\begin{equation}
\text{SCM} = 
\frac{\sum_{i=1}^{N} \left( X_a^i - \overline{X}_a \right) \left( X_b^i - \overline{X}_b \right)}
{\sqrt{\sum_{i=1}^{N} \left( X_a^i - \overline{X}_a\right)^2 \sum_{i=1}^{N} \left( X_b^i - \overline{X}_b \right)^2}}.
\end{equation}
SCM is effective for identifying changes in the relationships between frequency, since a high SCM value indicates that the spectral shapes are highly correlated.

Spectral Gradient Distance (SGD) captures changes in the spectral pattern by comparing the spectral gradient (GD). 
\begin{equation}
GD^i = X^{i+1} - X^i,
\end{equation}
\begin{equation}
\text{SGD} = \sqrt{\sum_{i=1}^{N-1} \left( GD_a^i - GD_b^i \right)^2}.
\end{equation}
SGD is helpful in identifying gradual or continuous changes in the spectral pattern.

Our approach is inspired by the Change Detection Metrics based on Spectral Shapes (DISS) proposed by \cite{yan2018novel}, which combines the SCM and SGD to assess changes in spectral shapes in remote sensing images. The DISS metric is defined as:
\begin{equation}
    \text{DISS} = \text{SGD} \cdot\left( \frac{1 - \text{SCM}}{2}\right).
\end{equation}
In this expression, the SGD is weighted by a factor $(1-\text{SCM})/2$; i.e., it adjusts the influence of SGD based on the Pearson correlation between spectral shapes. This weighting scheme reduces the sensitivity to amplitude changes caused by multiplicative factors. In situations where two time series segments have the same spectral shape but differ in overall amplitude, the SGD may detect high differences, but the SCM value is close to one. As a result, the weighting diminishes the impact of SGD in these cases, and makes the metric less responsive to uniform scaling that does not represent meaningful shape changes in the signal.


We adopt this idea and propose the Spectral Change Detection Metric (SCDM) by incorporating the SMC to create a more robust metric for detecting spectral changes. 
\begin{equation}
 \text{ SCDM}=  \text{SMC} +  \text{SGD} \cdot\left( \frac{1 - \text{SCM}}{2} \right).
\end{equation}
The inclusion of SMC allows the SCDM to capture changes in the average spectra, effectively detecting both abrupt and gradual changes in the signal while being robust to multiplicative factors. This makes it well-suited for detecting change points in our analysis of non-stationary time series data.

\begin{table*}[t!]
\captionsetup{justification=raggedright, singlelinecheck=false}
    \caption{Pattern Generation Equations and Parameter Values.}
    \label{Table:Gen_Equ}
    \centering
\begin{tabular}{@{}l l l l@{}}
\toprule
Pattern & Parameter & Generator Equation & Parameter Values \\ \midrule
Natural (NP) & - & $s_i(t) = \mu + r_i(t) \sigma$ & $\mu = 0, \sigma = 1$ \\
\addlinespace
Stratified (STP) & Random noise $(\sigma')$ & $s_{i}(t) = \begin{cases} \mu + \sigma'_1 r(t) & \text{if } t < \text{cp} \\ \mu + \sigma'_2 r(t) & \text{if } t \geq \text{cp} \end{cases}$ & $\sigma'_1 = 0.3\sigma, \sigma'_2 = 0.7\sigma$ \\
\addlinespace
Cycle (CP) & Amplitude $(\alpha),$ Period $(T)$ & $s_{i}(t) = \begin{cases} \mu + \sigma r(t) + \alpha_1 \sin\left(\frac{2 \pi t}{T_1}\right) & \text{if } t < \text{cp} \\ \mu + \sigma r(t) + \alpha_2 \sin\left(\frac{2 \pi t}{T_2}\right) & \text{if } t \geq \text{cp} \end{cases}$ & $\alpha_1 = 2\sigma, T_1=8, \alpha_2 = 4\sigma, T_2=16$ \\
\addlinespace
Trend (TP) & Slope $(s)$ & $s_{i}(t) = \begin{cases} \mu + \sigma r(t) + t \cdot s_1 & \text{if } t < \text{cp} \\ \mu + \sigma r(t) + t \cdot s_2 & \text{if } t \geq \text{cp} \end{cases}$ & $s_1 = 0.1\sigma, s_2 = 0.5\sigma$ \\
\addlinespace
Systematic (SYP) & Systematic departure $(d)$ & $s_{i}(t) = \begin{cases} \mu + \sigma r(t) + d_1 \cdot (-1)^t & \text{if } t < \text{cp} \\ \mu + \sigma r(t) - d_2 \cdot (-1)^t & \text{if } t \geq \text{cp} \end{cases}$ & $d_1 = 1\sigma, d_2 = 5\sigma$ \\
\addlinespace
Shift (SP) & Shift magnitude $(\eta)$ & $s_{i}(t) = \begin{cases} \mu + \sigma r(t) & \text{if } t < \text{cp} \\ \mu + \sigma r(t) + \eta & \text{if } t \geq \text{cp} \end{cases}$ & $\eta = 1\sigma$ \\
\bottomrule
\end{tabular}
\end{table*}
\subsubsection{Spectral Uncertainty Quantification}

In the final layer of the DGP, we employ MC sampling to quantify spectral uncertainty (SU) in our model predictions. The MC method is a computational algorithm that relies on repeated random sampling to approximate complex integrals and propagate uncertainties through mathematical models. 

To quantify SU, we generate $S$ sample paths ${ f^{(s)}(t) }_{s=1}^S$ from the posterior distribution of the DGP. Each sample function $f^{(s)}(t)$ represents a possible realization of the underlying process given the observed data and model uncertainties. Then we compute its absolute Fourier Transform to obtain the corresponding spectral representation based on Eq.~\ref{eq:STFT}, $\mathbf{F}^{(s)}(\omega) = \mathcal{F}| \mathbf{f}^{(s)}(t)|$. Then the mean $\mathbf{m}_F$ and variance $\mathbf{S}_F$ are:
\begin{equation}
    \mathbf{m}_F(\omega) = \frac{1}{S} \sum_{s=1}^S \mathbf{F}^{(s)}(\omega), 
\end{equation}
\begin{equation}
\mathbf{S}_F(\omega) = \frac{1}{S} \sum_{s=1}^S \left| \mathbf{F}^{(s)}(\omega) - \mathbf{m}_F(\omega) \right|^2,
\end{equation}
where $\mathbf{S}_F(\omega)$ is the direct measure of the SU. 


\subsubsection{Acquisition Function}

In order to determine subsequent input locations for sampling, we employ an AF, $a: X\rightarrow \mathbb{R}$, which integrates the spectral change metrics and spectral uncertainty: 
\begin{equation} 
a(x) = \beta \cdot \text{SCDM}(x) + (1 - \beta ) \cdot \text{SU}(x), 
\end{equation}
where $\beta $ is a hyperparameter that balances exploration and exploitation. The AF quantifies the desirability of evaluating a particular input location and guides the selection of the next sampling point $x^{\mathrm{next}}$. The next input is chosen by maximizing the AF such as
\begin{equation}
x^{\mathrm{next}}=\underset{x \in X}{\arg \max }\; a(x).
\end{equation}

By incorporating spectral change metrics and spectral uncertainty into the AL framework, we can achieve a balanced interplay between exploration, selecting data points from regions of high uncertainty, and exploitation, refining the model's understanding of change points based on the most informative data. This procedure is iteratively performed until the allocated budget is exhausted. As AL proceeds and an increasing number of samples are collected, the uncertainty of DGP spectral variance gradually diminishes. It also helps the model to capture the change patterns of the time series more accurately.

\subsection{Estimation and Evaluation}

\begin{algorithm}[t]
\caption{Spectral-AL CPD}
\label{alg:MCPThreshold}
\renewcommand{\algorithmicrequire}{\textbf{Input:}}
\renewcommand{\algorithmicensure}{\textbf{Output:}}
\begin{algorithmic}[1]
    \REQUIRE  \hspace*{\algorithmicindent}{}\\
   Set of sample points from AL loop: $ \mathcal{D} $\\ 
         Detection threshold:$ b$\\
        Suppression interval: $ \delta$\\
    \ENSURE Change points set: $ \mathcal{C}$
    \STATE Initialize $ \mathcal{C} \gets \emptyset $
    \WHILE{there exists $ x \in X $ such that $ \text{SCDM}(x) > b $}
        \STATE $ \hat{x}_k^{cp} \gets \arg \max_{x \in X} \; \text{SCDM}(x) $
        \STATE Add $ \hat{x}^{cp}_k $ to $\mathcal{C}$
        \STATE $ \text{SCDM}(x) \gets 0$ for all $ x \in (\hat{x}^{cp}_k - \delta, \hat{x}^{cp}_k + \delta) $
    \ENDWHILE
    \RETURN $ \mathcal{C} $
\end{algorithmic}
\end{algorithm}

The estimation procedure is outlined in Algorithm~\ref{alg:MCPThreshold}. Change points are identified based on a user-defined detection threshold $ b $. The algorithm iteratively detects change points $\hat{x}^{cp}_k$ by selecting the location with the highest SCDM value exceeding $ b$, suppressing the surrounding region $(\hat{x}^{cp}_k - \delta, \hat{x}^{cp}_k + \delta)$ to avoid multiple detections of the same change area, and repeating the process until no SCDM values exceed the threshold. 

For simulation experiments, we know the number of change points, the accuracy of the estimated change points is evaluated using the Root Mean Square Error (RMSE), which quantifies the average deviation between the detected change points and the ground truth:
    \begin{equation}
        \text{RMSE} = \sqrt{\frac{1}{K} \sum_{k=1}^{K} (\hat{x}^{cp}_k - x^{cp}_k)^2}
    \end{equation}
    where $ K$ is the number of true change points.

In cases where the number of change points is unknown, the F1 score is employed to assess performance \citep{van2020evaluation}. Let $ \mathcal{C} = \{\hat{x}^{cp}_1, \hat{x}^{cp}_2, \dots, \hat{x}^{cp}_L\} $ denote the set of change point locations identified by the detection algorithm, and let $\mathcal{T} = \{x^{cp}_1, x^{cp}_2, \dots, x^{cp}_K\}$ represent the set of ground truth change points. The set of true positives $\text{TP}(\mathcal{T}, \mathcal{C}) $ consists of those $ x^{cp}_k \in \mathcal{T} $ for which there exists $ \hat{x}^{cp}_l \in \mathcal{C} $ such that $ |x^{cp}_k - \hat{x}^{cp}_l| \leq M $, ensuring that each $ \hat{x}^{cp}_l \in \mathcal{C} $ is matched to at most one $ x^{cp}_k \in \mathcal{T} $ to avoid double counting.

The precision $ P $ and recall $ R $ are defined as:
\begin{align}
    P &= \frac{|\text{TP}(\mathcal{T}, \mathcal{C})|}{|\mathcal{C}|}, \\
    R &= \frac{1}{K} \sum_{k=1}^{K}\frac{|\text{TP}(\mathcal{T}_k, \mathcal{C})|}{|\mathcal{T}_k|}.
\end{align}
The F1 score is then calculated as:
\begin{equation}
    \text{F1 Score} = 2 \cdot \frac{P \cdot R}{P + R}.
\end{equation}
Here, the margin of error $ M $ is set to be equal to the suppression interval $ \delta $.

\section{Experiments}
\label{Sec:EXPERIMENTS}

\subsection{Simulation Experiments}
\begin{table*}[t]
  \caption{RMSE Comparison: 15 patterns evaluated under different explore-exploit parameters and the DACD baseline. Results are averaged over 10 simulations for single change-point detection. The best-performing method in each scenario is underlined.}
  \label{Table:RMSE}
  \centering
  \resizebox{\textwidth}{!}{%
  \begin{tabular}{@{}l|ccc|ccc|ccc|ccc|ccc|c@{}}
  \toprule
  Pattern Name & \multicolumn{3}{c}{AF($\beta=0$)} & \multicolumn{3}{c}{AF($\beta=0.25$)} & \multicolumn{3}{c}{AF($\beta=0.5$)} & \multicolumn{3}{c}{AF($\beta=0.75$)} & \multicolumn{3}{c|}{AF($\beta=1$)} & DACD \\ 
  \cmidrule(lr){2-4} \cmidrule(lr){5-7} \cmidrule(lr){8-10} \cmidrule(lr){11-13} \cmidrule(lr){14-16}
   & M-5 & M-15 & R-15 & M-5 & M-15 & R-15 & M-5 & M-15 & R-15 & M-5 & M-15 & R-15 & M-5 & M-15 & R-15 & \\ 
  \midrule
  Stratified (STP)  & 7.0 & 5.3 & 16.3 & 9.8 & 7.1 & 11.8 & 12.9 & 4.6 &  \underline{3.6} & 6.6 & 5.0 & 6.8 & 6.0 & 6.6 & 5.1 & 15.3 \\
  Cycle (CP) & 7.8 &  3.3 & 7.3 & 8.2 & 3.5 & \underline{3.1} & 8.9 & 4.1 &  5.2 & 5.4 &  4.4 & 3.9 &  2.8 & 4.9 & 6.2 & 22.4 \\
  Trend (TP)  & 2.1 &  1.0 & 0.7 & 3.2 &  0.8 & 4.4 & 5.1 &  \underline{0.4} & 1.6 & 4.6 &  0.7 & 2.1 & 7.1 &  0.9 & 0.8 & 0.5 \\
  Systematic (SYP) & 13.9 &  3.7 & 2.1 & 12.7 & 4.8 &  3.7 & 11.6 & 9.2 &  3.6 & 11.3 &  6.5 & 4.4 & 11.1 &  5.2 & \underline{1.1} & 21.3 \\
  Shift (SP)  & 0.4 &  \underline{0.0} & \underline{0.0} & 0.5 & 0.4 & \underline{0.0} & 0.9 &  0.3 & \underline{0.0} & 1.0 &  \underline{0.0} & \underline{0.0} & 0.9 &  0.3 & 0.3 & 21.5 \\
  Stratified-Cycle (STP-CP) & 11.4 &  7.0 & 8.8 & 8.9 &  6.5 & 4.0 & 3.8 &  1.7 & \underline{1.3} & 2.9 &  4.2 & 5.3 & 2.8 &  4.6 & 4.1 &  11.0 \\
  Stratified-Trend (STP-TP) & 0.5 &  0.7 & 0.4 & 0.9 &  0.7 & \underline{0.0} & 0.8 &  0.3 & 0.7 & 1.3 &  0.4 & 0.8 & 5.9 &  0.6 & 0.8 & 0.5 \\
  Stratified-Systematic (STP-SYP) & 12.1 & 13.8 &  8.8 & 9.6 & 7.8 &  7.2 & 7.6 & 4.8 &  4.5 & 10.8 & 13.5 &  4.8 & 11.1 &  \underline{3.1} & 7.3 & 23.5 \\
  Stratified-Shift (STP-SP) & 0.4 &  \underline{0.0} & \underline{0.0} &  0.5 & \underline{0.0} & \underline{0.0} & 0.6 &  \underline{0.0} & 0.1 & 0.6 &  \underline{0.0} & \underline{0.0} & 1.7 &  0.1 & 0.8 & 21.9 \\
  Cycle-Trend (CP-TP) &  1.0 & 0.5 & 1.0 & 0.8 &  \underline{0.4} & 1.6 & 1.1 & 1.0 &  0.8 & 3.0 &  0.7 & 0.8 &  1.0 & 0.5 & 0.9 & 2.0 \\
  Cycle-Systematic (CP-SYP) & 14.4 & 20.2 &  \underline{10.9} & 13.6 & 21.0 &  11.2 & 12.5 & 18.4 &  19.5 & 14.1 &  13.9 & 15.4 & 13.1 &  11.6 & 13.3 & 15.9 \\
  Cycle-Shift (CP-SP) & 0.9 &  \underline{0.3} & 1.0 & 0.8 & 1.0 &  0.4 & 1.0 &  0.5 & 1.4 & 0.7 & 0.7 &  0.5 & 0.9 &  0.5 & 1.1 & 20.9 \\
  Trend-Systematic (TP-SYP) & 1.0 &  0.4 & \underline{0.0} & 1.1 &  0.8 & 0.3 & 1.0 &  \underline{0.0} & 0.2 & 0.9 &  \underline{0.0} & 0.9 & 1.2 &  0.6 & 0.5 & 14.5 \\
  Trend-Shift (TP-SP) & 1.3 &  0.3 & 0.4 & 1.1 &  0.1 & 0.9 & 0.9 &  \underline{0.2} & 0.5 & 1.0 &  0.8 & 0.7 & 0.9 &  0.7 & \underline{0.2} & 4.0 \\
  Systematic-Shift (SYP-SP) &  \underline{0.0} & 0.9 & \underline{0.0} & 0.1 & 1.1 &  0.1 & 0.1 & 0.7 &  \underline{0.0} & \underline{0.0} & 1.0 &  \underline{0.0}&  \underline{0.0} & 1.0 & \underline{0.0} & 16.6 \\
  \bottomrule
  \end{tabular}%
  }
\vspace{0.3em}
\caption*{\textit{*M-5 refers to the Matérn kernel with window size $ A=5 $, M-15 corresponds to $ A=15 $, and R-5 represents the RBF kernel with $ A=15 $.}}
\end{table*}


\subsubsection{Generated Data} 

The simulation experiments contain both simple and complex patterns to represent a variety of time series behaviors. Each pattern consists of 100 data points with the time variable $t$ ranging from 0 to 100. A change point is introduced at $t=50$ to split the data into two different segments. Complex patterns are created by combining five basic patterns with different parameter choices, for a total of 15 patterns available for analysis. Table~\ref{Table:Gen_Equ} summarizes the equations and parameters used in the pattern generation process. Also see Figure~\ref{Fig:Change Patterns} for a visualization of these patterns.

The basic time series patterns include  
\begin{enumerate}
   \item Natural Pattern (NP): The NP is generated from a normally distributed process, where $s(t)$ represents the observation at time $t$, and $r_i(t)$ denotes white Gaussian noise. This pattern acts as a control case, reflecting a stable system with natural variability.
    \item Stratified Pattern (STP): The STP introduces a change in the variance of the system at a specified change point. 
    \item Cycle Pattern (CP): The CP models periodic oscillations in the system, with different amplitudes and periods before and after the change point. 
    \item Trend Pattern (TP): The TP introduces a linear trend to the system, which becomes steeper after the change point.     
    \item Systematic Pattern (SYP): The SYP introduces an alternating signal that systematically departs from the mean.
    \item Shift Pattern (SP): The SP involves a sudden change in the mean level of the system after the change point.  
\end{enumerate}


\subsubsection{Results}

We conducted experiments by selecting 10 points uniformly as the initial dataset and performing the AL loop for 10 iterations. The model was trained using a two-layer architecture with Matérn 5/2 and RBF kernels and optimized through the Adam optimizer with a learning rate of 0.1. For time-frequency translation, we used window sizes of $A=5$ and $A=15$. To evaluate our method, we compared it with the DACD approach \citep{zhao2023active}, which also uses AL for sample addition (see Appedix~\ref{Appendix:DACD}). Unlike our approach, DACD models the time series using a GP with an RBF derivative kernel and selects new samples by identifying locations with the highest mean derivative and uncertainty. Both methods employed the same initial datasets, AL iterations, and learning rate settings. Moreover, each experiment was repeated 10 times to reduce variability. The RMSE results for fifteen patterns under varying explore-exploit parameters $\beta$, as well as different kernels and window sizes, along with the DACD baseline results, are provided in Table~\ref{Table:RMSE}. 

From Table~\ref{Table:RMSE}, our method's performance varies with different $\beta$ values across the patterns. This variation emphasizes the importance of balancing exploration and exploitation in the AL process. Patterns with sharp shifts or simpler transitions (TP-SP) benefit more from larger $\beta$ values, where exploitation dominates. This strategy allows the model to refine estimates in regions likely to contain change points. In contrast, change patterns characterized by higher variability, such as CP, CP-TP, CP-SP, and CP-SYP, are more challenging to locate change points. These patterns achieve better results with lower $\beta$ values that favor exploration. Broader sampling helps the model detect change points that might be missed by a more focused strategy. Some patterns, like STP, obtain optimal results with intermediate $\beta$ values, which indicates that a balanced approach is necessary due to their complex dynamics. Additionally, patterns such as SYP-SP and STP-SP maintain lower RMSE values across different $\beta$ settings. This stability implies that these patterns are less sensitive to the sampling strategy, likely due to more straightforward change dynamics that are easier for the model to capture. 

Furthermore, the Matérn kernel with window size $A=15$ consistently achieves high accuracy across multiple patterns. While the RBF kernel also demonstrates competitive results in certain cases, it often struggles with mixed structures or more complex transitions. Our method consistently outperforms DACD in RMSE across all patterns. DACD is specifically designed to detect abrupt mean shifts and uses an RBF derivative kernel. However, with only 20 data points, DACD cannot accurately model the true distribution of the underlying processes in these diverse patterns, which leads to poor performance. In contrast, our spectral approach is more flexible and can capture complex dynamics more accurately. 

\begin{table*}[th!]
  \caption{F1 scores for real-world datasets.}
  \label{Table:F1}
  \centering
  \begin{tabular}{@{}l|cc|cc|cc|cc|cc@{}}
  \toprule
  Methods 
    & \multicolumn{2}{c|}{Occupancy}
    & \multicolumn{2}{c|}{Apple}
    & \multicolumn{2}{c|}{Run log}
    & \multicolumn{2}{c|}{Bee dance}
    & \multicolumn{2}{c}{Well log} \\
  \cmidrule(lr){2-3} \cmidrule(lr){4-5} \cmidrule(lr){6-7} \cmidrule(lr){8-9} \cmidrule(lr){10-11}
   & Matérn & RBF 
   & Matérn & RBF 
   & Matérn & RBF 
   & Matérn & RBF 
   & Matérn & RBF  \\
  \midrule
  DGP–Spectral–AF 
    &        &      
    &        &      
    &        &      
    &        &      
    &        &       \\
  $\beta=0$ 
    & 0.7142 & 0.5000 
    & 0.4286 & 0.4705     
    & 0.7500 & 0.7778       
    & 0.3076 & 0.2667 
    & 0.6923 & 0.3076 \\
  $\beta=0.25$ 
    & \underline{0.8235} & 0.6250 
    & \underline{0.7778} & 0.6316       
    & 0.8235 & \underline{0.8889}       
    & \underline{0.6667} & 0.3076 
    & \underline{0.8571} & 0.6250 \\
  $\beta=0.5$ 
    & 0.6667 & 0.5714 
    & 0.6250 & 0.5882 
    & 0.7142 & 0.8750 
    & 0.3636 & 0.2857 
    & 0.2857 & 0.5333 \\
  $\beta=0.75$ 
    & 0.7778 & 0.7368 
    & 0.5000 & 0.5556 
    & \underline{0.8889} & 0.7143 
    & 0.3077 & 0.3333 
    & 0.7692 & 0.4286 \\
  $\beta=1$ 
    & 0.7058 & 0.6154 
    & 0.5556 & 0.4286 
    & 0.7058 & 0.7058 
    & 0.2857 & 0.1428 
    & 0.5556 & 0.1667 \\
  DGP–Spectral–Random sample 
    & 0.3636 & 0.1428 
    & 0.2857 & 0.1538 
    & 0.4286 & 0.3077  
    & 0.1428 & 0.1538 
    & 0.5714 & 0.5926 \\
  DGP–Spectral–Uniform sample 
    & 0.5714 & 0.5882 
    & 0.1667 & 0.2667 
    & 0.5217 & 0.6153 
    & 0.2353 & 0.2857 
    & 0.6087 & 0.6400 \\
  \midrule
  ACPD~\citep{hayashi2019active} 
    & \multicolumn{2}{c|}{0.2857}
    & \multicolumn{2}{c|}{0.2143}
    & \multicolumn{2}{c|}{0.2759}
    & \multicolumn{2}{c|}{0.1818}
    & \multicolumn{2}{c}{0.4828} \\
  BOCPD~\citep{adams2007bayesian} 
    & \multicolumn{2}{c|}{0.8070}
    & \multicolumn{2}{c|}{0.5130}
    & \multicolumn{2}{c|}{\underline{1.000}}
    & \multicolumn{2}{c|}{0.1210}
    & \multicolumn{2}{c}{0.7960} \\
  ECP~\citep{matteson2014nonparametric} 
    & \multicolumn{2}{c|}{0.7500}
    & \multicolumn{2}{c|}{0.5130}
    & \multicolumn{2}{c|}{0.7920}
    & \multicolumn{2}{c|}{0.1240}
    & \multicolumn{2}{c}{0.8180} \\
  KCPA~\citep{harchaoui2008kernel} 
    & \multicolumn{2}{c|}{0.1070}
    & \multicolumn{2}{c|}{0.0290}
    & \multicolumn{2}{c|}{0.1390}
    & \multicolumn{2}{c|}{0.0100}
    & \multicolumn{2}{c}{0.0690} \\
  RBOCPDMS~\citep{knoblauch2018doubly} 
    & \multicolumn{2}{c|}{0.3290}
    & \multicolumn{2}{c|}{0.2390}
    & \multicolumn{2}{c|}{0.6600}
    & \multicolumn{2}{c|}{0.1790}
    & \multicolumn{2}{c}{0.5170} \\
  \bottomrule
  \end{tabular}
\end{table*}


\subsection{Real-World Experiments}

\subsubsection{Datasets}

\paragraph{Occupancy} Room occupancy measurements (temperature, humidity, light, $CO_2$) from the UCI Repository, downsampled every 16th observation. Includes 509 points, four dimensions, and eight change points. \footnote{All datasets are available in \cite{van2020evaluation}}

\paragraph{Apple} Daily closing price and volume of Apple Inc. stock from Yahoo Finance around 2000. Total of 622 points, two dimensions, and one notable change point on September 29, 2000.

\paragraph{Run log} Runner’s pace and total distance during interval training (running/walking), comprising 376 points, two dimensions, and eight change points corresponding to interval transitions.

\paragraph{Bee dance} Records a honeybee performing a three-stage waggle dance (left turn, right turn, waggle). Captured in four dimensions (x, y positions; sine, cosine of head angle), totaling 609 points with two change points.

\paragraph{Well log} Nuclear magnetic resonance measurements from drilling operations. Sampled every six iterations, resulting in 675 points and ten change points corresponding to shifts in rock stratification.

\begin{figure}[t]
  \centering
  \begin{subfigure}[b]{1.0\linewidth}
    \includegraphics[width=\linewidth]{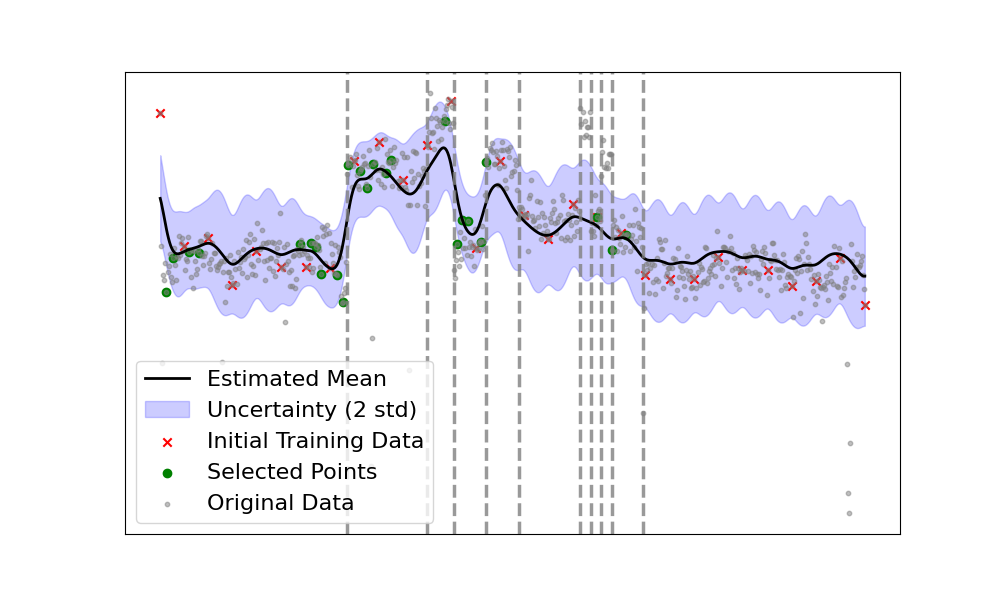}
      \vspace{-.3in}
     \caption{AL Iteration 5}
  \end{subfigure}
  \begin{subfigure}[b]{1.0\linewidth}
    \includegraphics[width=\linewidth]{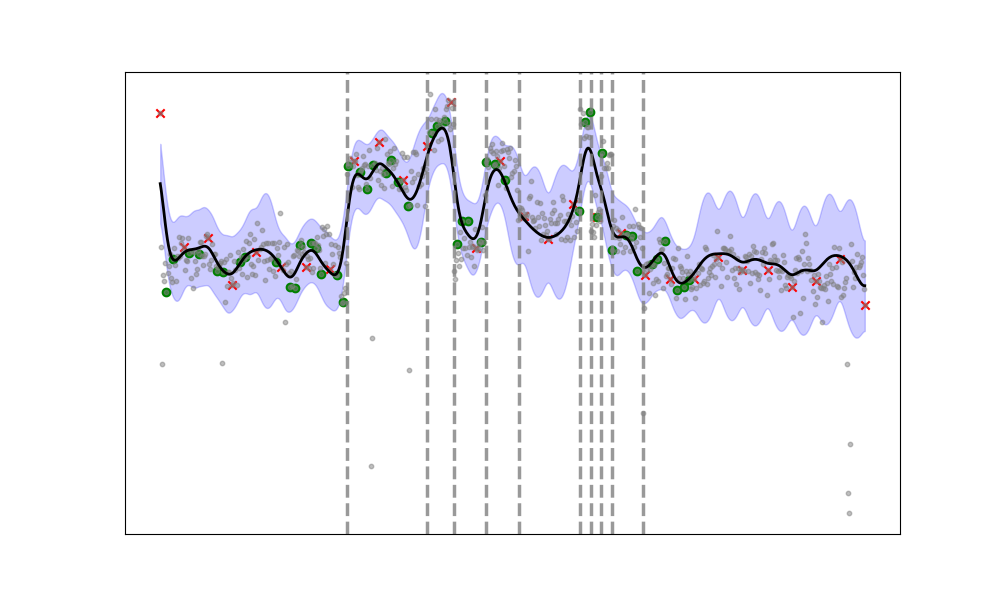}
          \vspace{-.3in}
    \caption{AL Iteration 10}
  \end{subfigure}
 \caption{DGP mean and variance predictions for the Well log dataset with an AF explore-exploit parameter $\beta$ = 0.75. Red crosses represent the initial training points, while green circles are the points selected during AL iterations 5 and 10, as shown in (a) and (b). Each iteration uses a batch size of 5.}
  \Description{Well log AL selected points.}
  \label{Fig:AL}
\end{figure}

\subsubsection{Evaluation Results}

We report results for two kernels at various $\beta$ values in Table~\ref{Table:F1}. The DGP was trained with a two-layer architecture using Matérn 5/2 and RBF kernels and optimized with the Adam optimizer (learning rate 0.01). We started with 30 uniformly sampled points as the initial dataset, then ran an AL loop for 10 iterations with a selection batch size of 5. For both datasets, window sizes and estimated distances were chosen based on domain knowledge and sampling frequency; specifically, we set $A=20$, $\delta=20$ for the Apple and Bee dance data, and $A=10$, $\delta=10$ for the Occupancy, Run log, and Well log data. We also compared our method with random and uniform sampling using the same data sizes. Notably, our method was trained on a limited dataset, whereas the baseline methods were trained on the full dataset (results reported in \cite{van2020evaluation}). Figure~\ref{Fig:AL} shows the AL training process of the Well log dataset at iterations 5 and 10, where the selected points concentrate around the true change points, which demonstrates the effectiveness of our proposed AF.

\subsubsection{Parameter Analysis}

\begin{table}[ht]
\centering
\caption{Sensitivity Analysis Results for Well log and Bee Dance Datasets}
\renewcommand{\arraystretch}{1.2}
\begin{tabular}{cc}
\begin{tabular}{@{}cccc@{}}
\toprule
\multicolumn{4}{c}{\textbf{Well log}} \\
$A$ & F1 ($\delta{=}10$) & $\delta$ & F1 ($A{=}10$) \\
\midrule
8   & 0.6925 & 4  & 0.5000 \\
10  & 0.8571 & 6  & 0.5323 \\
12  & 0.7500 & 8  & 0.7500 \\
14  & 0.6154 & 10 & 0.8571 \\
16  & 0.5333 & 12 & 0.8571 \\
\bottomrule
\end{tabular}
&
\begin{tabular}{@{}cccc@{}}
\toprule
\multicolumn{4}{c}{\textbf{Bee Dance}} \\
$A$ & F1 ($\delta{=}20$) & $\delta$ & F1 ($A{=}20$) \\
\midrule
10  & 0.2857 & 10 & 0.6667 \\
15  & 0.4000 & 15 & 0.6667 \\
20  & 0.6667 & 20 & 0.6667 \\
25  & 0.0000 & 25 & 0.6667 \\
30  & 0.0000 & 30 & 0.6667 \\
\bottomrule
\end{tabular}
\end{tabular}
\label{table:sensitivity}
\end{table}

We conducted sensitivity analysis on two real-world datasets (Well log and Bee Dance) by varying the spectral window size $A$ and the suppression interval $\delta$, as summarized in Table~\ref{table:sensitivity}. The window size $A$ controls the size of the sliding window used in the Fourier transform for spectral analysis. Smaller windows are sensitive to local changes but may generate false positives due to noise. Larger windows are more robust to noise but may miss subtle, localized changes. The suppression interval $\delta$ is used to avoid multiple detections near the same change region. A small $\delta$ may lead to redundant detections around a single change point, while a large $\delta$ may cause closely spaced changes to be missed.
We can examine the spectral metric to guide the selection of $\delta$ and set natural thresholds $b$, for instance, significant peaks can be identified using a percentile-based threshold (e.g., top 5\%). The distribution of peak locations also informs the choice of $\delta$. Closely grouped detections may indicate the need for a larger $\delta$, while sparse peaks may suggest lowering it or adjusting the threshold.

\section{Conclusion}
\label{Sec:CONCLUSION}

In summary, this study presents a novel CPD algorithm that integrates DGPs, spectral analysis, and AL strategies to detect various change patterns. The incorporation of DGPs provides a flexible modeling framework capable of capturing complex data structures, while AL enhances detection efficiency by sequentially selecting the most informative data points for analysis. By leveraging the adaptability of spectral metrics, our method can handle a wide range of time series patterns without making explicit assumptions about the data generating mechanism. Through multiple simulations and real-world experiments, we demonstrate that our approach effectively detects change points across diverse data patterns.

\bibliographystyle{ACM-Reference-Format}

\appendix

\section{Baseline Methods}
\label{Appendix:DACD}

\paragraph{DACD}
The DACD \citep{zhao2023active} approach is designed for CPD that incorporates GP derivatives into the AL framework. The method works by fitting a GP model to an initial dataset to adapt to the underlying structure of the process. DACD utilizes derivative GP information to identify abrupt changes through maxima in the first-order derivative of the process. The GP derivative process is then used to formulate the AF in AL, which selects the most informative data points for detecting change points. For instance, the Expected Improvement (EI) AF is defined as:
\[
a_{\mathrm{EI}}(x) = \sigma^\nabla(x)(\gamma(x)\Psi(\gamma(x))+ \Phi(\gamma(x))),
\]
where $\gamma(x)=\frac{\mu^\nabla(x)-f^\nabla(x^+)-\xi}{\sigma^\nabla(x)}$,  $\mu^\nabla(x)$ and $\sigma^\nabla(x)= \sqrt{k^\nabla\left(x,x\right)}$ represent the mean and variance of the GP derivative, while the hyperparameter $\xi$ modulates the balance between exploration and exploitation.  $\Psi$ and $\Phi$ are the cumulative distribution function and density function of a standard Gaussian distribution, respectively. 

\paragraph{ACPD} Active Change-Point Detection (ACPD) \citep{hayashi2019active} is an unsupervised method for detecting change points by actively selecting the most informative time points. It starts with a few initial samples and computes their change scores. A GP is used to model the score function, and an acquisition function like EI is used to choose new points to sample. After a fixed number of AL steps, the final change points are selected using a segmentation-based method with a Bayesian information criterion (BIC) penalty.
\end{document}